\documentclass[runningheads,a4paper]{llncs}

\usepackage{amssymb}
\usepackage{graphicx}

\usepackage{times}
\usepackage{graphicx} 
\usepackage[caption=false]{subfig}

\usepackage{algorithm}
\usepackage{algorithmic}

\usepackage{hyperref}

\usepackage{tabularx}
\usepackage{amssymb}
\usepackage{amsmath}
\usepackage{bm}

\DeclareMathOperator*{\argmax}{argmax}

\newcommand{\x}{\mathbf{x}}

\newcommand{\DD}{\mathbf{D}}
\newcommand{\I}{\mathbf{I}}

\newcommand{\N}{\mathcal{N}}

\newcolumntype{C}{>{\centering\arraybackslash}X}

\usepackage{url}
\urldef{\mailsa}\path|{srijith, balamurugan, shirish}@iisc.ernet.in|    
\newcommand{\keywords}[1]{\par\addvspace\baselineskip
\noindent\keywordname\enspace\ignorespaces#1}

\begin{document}

\mainmatter  

\title{Gaussian Process Pseudo-Likelihood Models for  Sequence Labeling}

\titlerunning{Gaussian Process Pseudo-Likelihood Models for  Sequence Labeling}

%
%
\author{P. K. Srijith\inst{1} \and P. Balamurugan\inst{2} \and Shirish Shevade\inst{3} }
\authorrunning{P. K. Srijith \and P. Balamurugan \and Shirish Shevade}

\institute{Department of Computer Science, University of Sheffield,  United Kingdom \\ \email{pk.srijith@dcs.shef.ac.uk} \and SIERRA Project Team, INRIA-ENS, Paris, France \\ \email{balamurugan.palaniappan@inria.fr} \and Computer Science and Automation, Indian Institute of Science, Bangalore \\
\email{shirish@csa.iisc.ernet.in}
}

%
%

\toctitle{Gaussian Process Pseudo-Likelihood Models for  Sequence Labeling}
\maketitle

\begin{abstract}

Several machine learning problems arising in  natural language processing can be modeled as a sequence labeling problem.  Gaussian processes (GPs) provide a Bayesian approach to learning such problems in a kernel based framework. We develop Gaussian process models based on  pseudo-likelihood to solve sequence labeling problems.  The  pseudo-likelihood model enables one to capture multiple dependencies among the output components of the sequence without becoming computationally intractable.  We use an efficient variational Gaussian approximation method to perform inference in the proposed model. We also provide an iterative algorithm which can effectively make use of the information from the neighboring labels to perform prediction. The ability to capture multiple dependencies makes the proposed approach useful for a wide range of sequence labeling problems.  Numerical experiments on some sequence labeling problems in natural language processing demonstrate the usefulness of the proposed approach. \\
\keywords{Gaussian processes, sequence labeling,  variational inference}

\end{abstract}

\section{Introduction}

Sequence labeling is the task of classifying a sequence of inputs into a sequence of outputs. It arises commonly in natural language processing (NLP) tasks such as part-of-speech tagging, chunking, named entity recognition etc.  For instance, in part-of-speech (POS) tagging, the input is a  sentence and the output is a sequence of POS tags.  The output consists of components whose labels depend on the labels of other components in the output. Sequence labeling takes into account these inter-dependencies among various components of the output~\cite{lsp}.

In recent years, sequence labeling has received considerable attention from the machine learning community and is often studied under the general framework of structured prediction. Many algorithms have been proposed to tackle sequence labeling problems.  Hidden Markov model (HMM)~\cite{hmm},  conditional random field (CRF)~\cite{crf} and structural  support vector machine (SSVM)~\cite{structSVM} are the popular algorithms for sequence labeling. SSVM allows learning a SVM for predicting a structured output including sequences. It is based on a large margin framework and is not probabilistic in nature. HMM is a probabilistic directed graphical model based on Markov assumption and has been widely used for problems in speech and language processing. CRF is also a probabilistic model based on Markov random field assumption. These parametric approaches can provide an estimate of uncertainty in predictions due to their probabilistic nature. However, they do not follow a  Bayesian approach as they make a pointwise estimate of their parameters. This makes them less robust and heavily dependent on cross-validation for model selection. Bayesian CRF~\cite{bcrf} overcomes this problem by providing a Bayesian treatment to CRF. Approaches like SSVM and maximum margin Markov network (M3N)  make use of
kernel functions which overcome the limitations arising
due to the parametric nature of models such as CRF. Kernel CRF~\cite{kcrf} is proposed to overcome this limitation of the CRF,  but it is also not a Bayesian approach. 

Gaussian processes (GPs)~\cite{ras05} have emerged as a better alternative to offer a
non-parametric fully Bayesian approach to solve the sequence labeling problem. 
An initial work which studied Gaussian process for sequence labeling is  \cite{altun04}, where GPs were proposed as an alternative to overcome the limitations of CRF; however they used a maximum a posteriori (MAP) approach instead  of a fully Bayesian approach. This caused problems of model selection and robustness issues. A more recent work GPstruct~\cite{bratiers13} provides a Bayesian approach to general structured prediction problem with GPs.
It uses Markov Chain Monte Carlo (MCMC) method to obtain the  posterior  distribution which slows down the inference. Their approach is based on Markov  random field assumption which could not capture long range dependencies among the labels. 
This difficulty is  overcome in \cite{bratiers14} which uses an approximate likelihood to reduce the computational complexity arising due to the consideration of larger dependencies. In \cite{bratiers14}, the proposed model was used to solve grid structured problems in computer vision and was found to be effective in these 
problems.

In this work, we develop a Gaussian process approach  based on pseudo-likelihood to solve sequence labeling problems (which we call GPSL).  The GPSL model helps to capture multiple dependencies among the output components in a sequence without becoming computationally intractable. We develop a variational inference method to obtain the posterior which is faster than  MCMC based approaches and does not suffer from convergence problems. We also provide an efficient algorithm to perform prediction in the GPSL model which  effectively takes into account the dependence on multiple output components.  We consider various GPSL models which consider different number of dependencies. We study the usefulness of these models on various sequence labeling problems arising in natural language processing (NLP). The GPSL models which capture more dependencies are found to be useful for these sequence labeling problems. They are also useful in sequence labeling data sets where the labels might be missing for some output components, for example, when the labels are obtained using crowd-sourcing. The main contributions of the paper are as follows :
\begin{enumerate}
 \item A faster training algorithm based on variational inference.
 \item An efficient prediction algorithm which considers multiple dependencies. 
 \item Application to sequence labeling problems in NLP.
\end{enumerate}

The rest of the paper is organized as follows. Gaussian processes are introduced in Section~\ref{backgpsp}. Section~\ref{gpsp} discusses the proposed approach, Gaussian process sequence labeling (GPSL), in detail.  We provide details of the variational inference and prediction algorithm for the GPSL model in Section~\ref{vigpsp} and Section~\ref{predgpsp} respectively.  In Section~\ref{exptgpsp}, we study the performance of various GPSL models on  sequence labeling problems and draw several conclusions in Section~\ref{congpsp}.

{\bf Notations: }
We consider a sequence labeling problem over sequences of input-output space pair ($\mathcal{X,Y}$). The  input sequence space $\mathcal{X}$ is assumed to be made up of $L$ components $\mathcal{X} = \mathcal{X}_1 \times \mathcal{X}_2 \times \ldots \mathcal{X}_L$  and the associated  output sequence space has $L$ components $\mathcal{Y} = \mathcal{Y}_1 \times \mathcal{Y}_2 \times \ldots \mathcal{Y}_L$.   We assume a one-to-one mapping between the input and output components. Each component of the output space is assumed to take a discrete value from the set  $\{1,2,\ldots, J\}$. Each component in the input space is assumed to belong to a $P$ dimensional space $\mathcal{R}^P$ representing features for that input
component. Consider a  collection of $N$ training input-output examples $\DD = \{(\mathbf{x}_n, \mathbf{y}_n)\}_{n=1}^N$, where each example $(\mathbf{x}_n, \mathbf{y}_n)$ is such that $\mathbf{x}_n \in \mathcal{X}$ and $\mathbf{y}_n \in \mathcal{Y}$. Thus, $\mathbf{x}_n$ consists of $L$ components $(\x_{n1}, \x_{n2}, \ldots, \x_{nL})$ and  $\mathbf{y}_n$ consists of $L$ components $(y_{n1}, y_{n2}, \ldots, y_{nL})$.  The training data $\DD$ contains $NL$ input-output components. 


\section{Background}
\label{backgpsp}
A Gaussian process (GP) is a collection of random variables with the property that the joint distribution of any finite subset of which is a Gaussian \cite{ras05}. It generalizes Gaussian distribution to infinitely many random variables and is  used as a prior over a latent  function.  The GP is completely specified by a mean function and a covariance function. The covariance function is defined over latent function values of a pair of inputs and is evaluated using the Mercer kernel function over the pair of inputs. The covariance function expresses some general properties of functions such as their smoothness, and length-scale.  A commonly used covariance function is the squared exponential (SE) or the Gaussian kernel
\begin{eqnarray}
\label{eqn:covfun}
 cov\bigl(f(\x_{mi}),f(\x_{nl})\bigr)  =  K(\x_{mi},\x_{nl}) 
=  \sigma_f^2 \exp(-\frac{\kappa}{2}||\x_{mi} - \x_{nl}||^2).
\end{eqnarray}
Here $f(\x_{mi})$ and $f(\x_{nl})$ are latent function values associated with the input components $\x_{mi}$ and $\x_{nl}$ respectively. $\bm{\theta}  = (\sigma_f^2, \kappa)$ denotes the hyper parameters associated with the covariance function $K$.

Multi-class classification approaches are useful when the output consists of a single component taking values from a finite discrete set  $\{1,2,\ldots, J\}$.  Gaussian process multi-class classification approaches~\cite{williams98,girolami06,chai12} associate a latent function $f^{j}$ with every label $j\in \{1,2,\ldots, J\}$.
Let the vector of latent function values associated with a particular label $j$  over all the training examples be $\mathbf{f^j}$.
The latent function $f^j$  is assigned an independent GP prior  with zero mean and covariance function $K^j$ with hyper parameters $\bm{\theta}_j$ . Thus, $\mathbf{f^j} \sim N(0,\mathbf{K^j})$, where $\mathbf{K^j}$ is a matrix obtained by evaluating the covariance function $K^j$ over all the pairs of training data input components.

In  multi-class classification, the likelihood  over a multi-class output $y_{nl}$ for an input $\x_{nl}$  given the latent functions is defined as~\cite{ras05} 
\begin{equation}
\label{GPClikelihood}
p(y_{nl}|f^1(\x_{nl}), f^2(\x_{nl}), \ldots, f^J(\x_{nl})) = \frac{\exp(f^{y_{nl}}(\x_{nl}))}{\sum_{j=1}^J f^j(x_{nl})}.
\end{equation}
The likelihood (\ref{GPClikelihood}) is known as multinomial logistic or softmax function and is used widely for the GP multi-class classification problems~\cite{williams98,chai12}. It is important to note that the likelihood function (\ref{GPClikelihood}) used for the multi-class classification problems is not  Gaussian. Hence, the posterior over the latent functions cannot be obtained in a closed form. GP multi-class classification approaches work by approximating the posterior as a Gaussian using approximate inference techniques such as  Laplace approximation~\cite{williams98} and variational inference~\cite{girolami06,chai12}. The Gaussian approximated posterior is then used to make predictions on the test data points. These approximations also yield an approximate marginal likelihood or a lower bound on marginal likelihood which can be used to perform model selection~\cite{ras05}.

A sequence labeling problem can be treated as a multi-class classification problem.  One can use multi-class classification to obtain a label for each component of the output independently. But this fails to take into account the inter-dependence among components. If one considers the entire output as a distinct class, then there would be an exponential number of classes and the learning problem becomes intractable.  Hence, the sequence labeling problem has to be studied separately from the multi-class classification problems.

\section{Gaussian Process Sequence Labeling}
\label{gpsp}

Most of the previous approaches~\cite{crf,bratiers13}  to sequence labeling use  likelihood based on Markov random field assumption which captures only the interaction between neighboring output components. Non-neighboring components also play a significant role in problems such as sequence labeling. In these models, capturing such interactions are computationally expensive due to large clique size.  The proposed approach,  Gaussian process sequence labeling (GPSL),  can take into account interactions among various output components  without becoming computationally intractable by using a pseudo-likelihood (PL) model~\cite{besag1975}. 

The PL model defines the likelihood of an output $\mathbf{y_n}$ given the input $\mathbf{\x_n}$ as $p(\mathbf{y_n}|\mathbf{\x_n})$ $ \propto \prod_{l = 1}^L p(y_{nl}|\x_{nl}, \mathbf{y_n}\backslash y_{nl})$.
where, $\mathbf{y_n}\backslash y_{nl}$ represents all labels in $\mathbf{y_n}$ except $y_{nl}$.
PL models have been successfully used  to address many sequence labeling problems in natural language processing \cite{toutanova03,sutton07}. They can capture long range dependencies without becoming computationally intractable as the normalization is done for each output component separately.
In models such as CRF, normalization is done over the entire output. This renders them incapable of  capturing long range dependencies as the  number of summations in the normalization grows exponentially. The PL model is different from a locally normalized model like maximum entropy Markov model (MEMM) as each output component depends on several other output components. Therefore, they do not suffer from the label bias problem~\cite{lsp} unlike MEMM.
However, PL models create cyclic dependencies among the output components~\cite{heckerman01} and this makes prediction hard. We discuss an efficient approach to perform prediction in this case in Section~\ref{predgpsp}.

\begin{figure*}[t]
\vskip 0.2in
\begin{center}
\subfloat[ Dependence among input and output components. Dependence on various output components are modelled separately.]{
\includegraphics[scale=0.45]{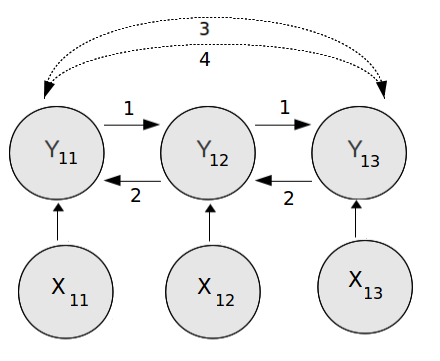}
\label{fig.cdn1}
}
\quad
\subfloat[ Dependence of local and dependent latent functions. The local latent functions are defined over input-output pairs and dependent latent functions are defined between output components. ]{
\includegraphics[scale=0.47]{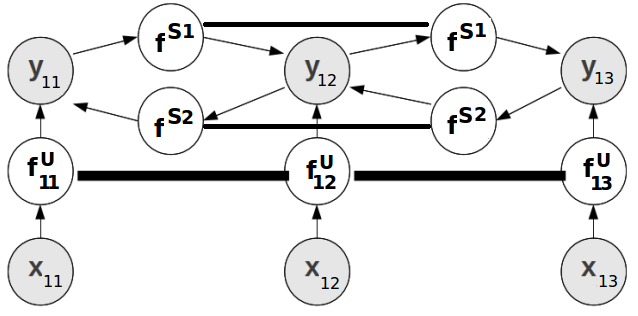}
\label{fig.cdn2}
}
\caption{Dependence of latent functions and input-output components in Gaussian process sequence labeling model.}
\label{fig.cdn}
\end{center}
\vskip -0.2in
\end{figure*}


The label of an output component need not depend on the labels of all the other output components. The dependencies among these output components are captured through the set $S$.
Consider the  directed graph in Figure~\ref{fig.cdn1} for a sequence labeling problem,  where each output component is assumed to depend only on the neighboring output components.
Here, the dependency set $S =  \{1,2\}$, where $1$ denotes the dependence of an output component on the previous output component and $2$ denotes its dependence on the next output component.
One can also consider a model where an output component depends on the previous two output components and the next two output components.
Let  $R$ denote the number of dependency relations in a set $S$ (that is, $R$ is the cardinality of $S$) and we assume it to be the same for all the output components for the sake of clarity in presentation.
Taking into account those dependencies, we can redefine the likelihood  as 
\begin{equation}
\label{eqn:redPL}
p(\mathbf{y_n}|\mathbf{\x_n}) \propto \prod_{l = 1}^L p(y_{nl}|\x_{nl}, \mathbf{y_{nl}^{S}}).
\end{equation}
Here, $\mathbf{y_{nl}^{S}}$ denotes the set of labels $\{y_{nl}^{d}\}_{d=1}^R$ of the output components referred by the dependency set $S$ and  $y_{nl}^{d}$ denotes the label of the $d^{th}$ dependent output component.   
In (\ref{eqn:redPL}), instead of conditioning on the rest of the labels,  we condition $y_{nl}$ only on the labels defined by the dependency set $S$.

Now, the likelihood $p(y_{nl}|\x_{nl}, \mathbf{y_{nl}^{S}})$ can be defined using a set of latent functions. We use different latent functions to model different dependencies. The dependency of the label $y_{nl}$ on $\x_{nl}$ is defined as a local dependency and is modeled as in GP multi-class classification. We associate a latent function with each label in the set $\{1,2,\ldots J\}$. The latent function associated with a label $j$, denoted as $f^{Uj}$, is called a local latent function. It is defined over all the training input components $\x_{nl}$ for every  $n$ and $l$ and the latent function values associated with a particular label $j$ over $NL$ training examples are denoted by $\mathbf{f^{Uj}}$. The local latent functions associated with a particular input component $\x_{nl}$ are denoted as $\mathbf{f^{U}_{nl}} = \{f^{U1}_{nl}, \ldots, f^{UJ}_{nl}\}$.  We also associate a latent function $f^{Sd}$ with each dependency relation $d \in S$ and call them dependent latent functions. These latent functions are defined
over all the values of a pair of labels $( \hat{y}_{nl}, y_{nl})$ where $\hat{y}_{nl} \in \{1,2,\ldots J\}$ and $y_{nl} \in \{1,2,\ldots J\}$.  The latent function values associated with a particular dependency $d$ over $J^2$ label pair values are denoted by $\mathbf{f^{Sd}}$. The dependence of various latent functions on the input and output components for the directed graph in Figure~\ref{fig.cdn1} is depicted in Figure~\ref{fig.cdn2}. Given these latent functions we define the likelihood $p(y_{nl}|\x_{nl},  \mathbf{y_{nl}^{S}})$ to be  a member of an exponential family: 
\begin{eqnarray}
\label{eqn:strGPlik}
& &\hspace{-4mm} p(y_{nl}|\x_{nl}, \mathbf{y_{nl}^{S}}, \{\mathbf{f^{Uj}}\}_{j=1}^J, \{\mathbf{f^{Sd}}\}_{d=1}^R) = \nonumber \\
& &\hspace{-4mm} \frac{\exp(f^{Uy_{nl}}(\x_{nl}) + \sum_{d=1}^R f^{Sd}(y^d_{nl},y_{nl}))}{\sum_{y_{nl} = 1}^J \exp(f^{Uy_{nl}}(\x_{nl}) + \sum_{d=1}^R f^{Sd}(y^d_{nl},y_{nl}))}.
\end{eqnarray}
This differs from the softmax likelihood (\ref{GPClikelihood}) used in  multi-class classification in that it captures the dependencies among output components. Given the latent functions and the input $\mathbf{X} = \{\mathbf{\x_n}\}_{n=1}^N$, the likelihood of the output $\mathbf{Y} = \{\mathbf{y_n}\}_{n=1}^N$ is 
\begin{eqnarray}
& &\hspace{-5mm} p(\mathbf{Y}|\mathbf{X},\{\mathbf{f^{Uj}}\}_{j=1}^J, \{\mathbf{f^{Sd}}\}_{d=1}^R) =  \prod_{n = 1}^N \prod_{l = 1}^L  p(y_{nl}|\x_{nl}, \mathbf{y_n}_{\{D_{nl}\}}, \{\mathbf{f^{Uj}}\}_{j=1}^J, \{\mathbf{f^{Sd}}\}_{d=1}^R)
\end{eqnarray}

We impose independent GP priors over the latent functions $\{f^{Uj}\}_{j=1}^J, \{f^{Sd}\}_{d=1}^R$. The latent function $f^{Uj}$ is given a zero mean GP prior with covariance function $K^{Uj}$ parameterized by $\bm{\bm{\theta}}_j$.  Thus, $\mathbf{f^{Uj}}$ is a Gaussian with mean $0$ and covariance $\mathbf{K^{Uj}}$ of size $NL \times NL$,  that is  $p(\mathbf{f^{Uj}}) = \N(\mathbf{f^{Uj}}; \mathbf 0, \mathbf{K^{Uj}})$.  $\mathbf{K^{Uj}}$ consists of covariance function evaluations over all the pairs of training data input components  $\{\{\x_{nl}\}_{l=1}^L\}_{n=1}^N$.  The latent function $f^{Sd}$ is given zero mean GP prior with an identity covariance which is defined to be $1$ when inputs are the same and $0$ otherwise. Thus $\mathbf{f^{Sd}}$ is a Gaussian with mean $0$ and covariance $\I$ of size $J^2$,  that is  $p(\mathbf{f^{Sd}}) = \N(\mathbf{f^{Sd}};\mathbf{0},\I_{J^2})$. Let $\mathbf{f^U} = (\mathbf{f^{U1}}, \mathbf{f^{U2}}, \ldots, \mathbf{f^{UJ}})$ be the collection of
all local latent functions and  $\mathbf{f^S} = (\mathbf{f^{S1}}, \mathbf{f^{S2}}, \ldots, \mathbf{f^{SR}})$ be the collection of all dependent latent functions. Then  the prior over $\mathbf{f^U}$ and $\mathbf{f^S}$  is defined as

\begin{equation}
\label{prior}
p(\mathbf{f^U,f^S|X})
=
\mathcal{N}\biggl( \begin{bmatrix}
   \mathbf{f^U}  \\
  \mathbf{f^S}
\end{bmatrix}; \mathbf{0}, \begin{bmatrix}
   \mathbf{K^U} &   0 \\
   0 &   \mathbf{K^S}
\end{bmatrix}\biggr),
\end{equation}
where $\mathbf{K^U} = diag(\mathbf{K^{U1}}, \mathbf{K^{U2}}, \ldots, \mathbf{K^{UJ}})$ is a block diagonal matrix  and $\mathbf{K^S} = \mathbf{I}_{J^2} \otimes \mathbf{I}_{R}$.

The posterior over the latent functions $p(\mathbf{f^U,f^S}|\DD)$ is
\begin{equation}
\label{eqn:pos}
p(\mathbf{f^U,f^S}|\mathbf{X,Y}) = \frac{1}{p(\mathbf{Y|X})} p(\mathbf{Y}|\mathbf{X}, \mathbf{f^U,f^S}) p(\mathbf{f^U,f^S|X}) \nonumber
\end{equation}
where $p(\mathbf{Y|X}) = \int p(\mathbf{Y}|\mathbf{X}, \mathbf{f^U,f^S}) p(\mathbf{f^U,f^S|X}) d\mathbf{f^U} d\mathbf{f^S}$ is called  evidence. Evidence is a function of hyper-parameters $\bm{\bm{\theta}} = (\bm{\bm{\theta}}_1, \bm{\bm{\theta}}_2, \ldots, \bm{\theta}_J)$ and is maximized to estimate them. For notational simplicity, we suppress the dependence of evidence, posterior and prior on the hyper-parameter $\bm{\theta}$.  Due to the non-Gaussian nature of the likelihood, evidence is intractable  and  the posterior cannot be determined exactly. We use a variational inference technique to obtain an approximate posterior. Variational inference is faster than sampling based techniques used in \cite{bratiers13}  and does not suffer from convergence problems \cite{pml}. It can easily handle multi-class problems and is scalable to models with a large number of parameters. Further, it provides an approximation to the evidence which is useful in estimating the hyper-parameters of the model.

\section{Variational Inference}
\label{vigpsp}

A variational Inference technique~\cite{pml} approximates the intractable posterior by an approximate variational distribution. It approximates the posterior $p(\mathbf{f|X,Y})$  by a variational distribution $q(\mathbf{f|\bm{\bm{\gamma}}})$, where $\mathbf{f} = (\mathbf{f^U,f^S})$ and $\mathbf{\bm{\gamma}}$ represents the variational parameters. In variational  inference, this is done by minimizing the Kullback-Leibler (KL) divergence between $q(\mathbf{f|\bm{\gamma}})$ and $p(\mathbf{f|X,Y})$. This is often intractable and the variational parameters are obtained by maximizing a variational lower bound $L(\bm{\theta}, \bm{\gamma})$.
\begin{eqnarray}
\label{eqn:vi}
& &  KL(q(\mathbf{f|\bm{\gamma}}) || p(\mathbf{f|X,Y}))  =  -L(\bm{\theta}, \bm{\gamma}) + \log p(\mathbf{Y|X}) \\
\label{eqn:lb}
& &  \mbox{where } L(\bm{\theta}, \bm{\gamma}) = -KL(q(\mathbf{f}|\bm{\gamma}) || p(\mathbf{f|X})) + 
 \int q(\mathbf{f}|\bm{\gamma})\log p(\mathbf{Y}|\mathbf{X}, \mathbf{f}) d\mathbf{f} . \nonumber
\end{eqnarray}
Maximizing the variational lower bound $L(\bm{\theta}, \bm{\gamma})$ results in minimizing the KL divergence $KL(q(\mathbf{f|\bm{\gamma}}) || p(\mathbf{f|X,Y}))$, since the evidence $p(\mathbf{Y|X})$ does not depend on the variational parameters.

We use a variational Gaussian (VG) approximate inference approach \cite{opper09} where the variational distribution is assumed to be  a Gaussian. Variational Gaussian approaches can be slow because of the requirement to estimate the covariance matrix. Fortunately, recent advances in VG inference approaches \cite{opper09} enable one to compute the covariance matrix using  $\mathcal{O}(NL)$ variational parameters. In fact, we use the VG approach for GPs~\cite{khan12} which requires  computation of only $\mathcal{O}(NL)$  variational parameters, but at the same time uses a concave variational lower bound. We assume the variational distribution $q(\mathbf{f}|\bm{\gamma})$ takes the form of a Gaussian distribution and factorizes as $q(\mathbf{f^U}|\bm{\gamma^U})q(\mathbf{f^S}|\bm{\gamma^U})$ where $\bm{\gamma} = \{\bm{\gamma^U}, \bm{\gamma^S} \}$. Let  $q(\mathbf{f^U}|\bm{\gamma^U}) = \N(\mathbf{f^U; m^U, V^U})$ where $\bm{\gamma^U} = \{\mathbf{m^U, V^U}\}$ and $q(\mathbf{f^S}) = \N(\mathbf{f^S; m^S, V^S})$
where $\bm{\gamma^S} = \{\mathbf{m^S, V^S}\}$. Then, the variational lower bound $L(\bm{\theta}, \bm{\gamma})$ can be written as
\begin{eqnarray}
& &\hspace{-2mm} L(\bm{\theta}, \bm{\gamma}) =  \frac{1}{2} ( \log|\mathbf{V^U\Omega^U}| + \log|\mathbf{V^S\Omega^S}| - tr(\mathbf{V^U\Omega^U})   - tr(\mathbf{V^S\Omega^S})    \\
& & \hspace{-2mm} -  \mathbf{{m^U}^\top\Omega^U m^U}   - \mathbf{{m^S}^\top\Omega^S m^S} ) +  \sum_{n=1}^{N} \sum_{l=1}^{L} \mathbb{E}_{q(\mathbf{f^U}|\bm{\gamma^U})q(\mathbf{f^S}|\bm{\gamma^S})} [\log p(y_{nl}|\x_{nl}, \mathbf{y_{nl}^S}, \mathbf{f})] \nonumber
\end{eqnarray}
where $\mathbf{\Omega^U} = \mathbf{{K^U}^{-1}}$, $\mathbf{\Omega^S} = \mathbf{{K^S}^{-1}}$ and $\mathbb{E}_{q(x)} [f(x)] = $ $\int f(x) q(x) dx$ represents the expectation of $f(x)$ with respect to the density $q(x)$. Since $\mathbf{K^U}$ is block diagonal, its inverse is block diagonal, and hence $\mathbf{\Omega^U}$ is block diagonal  that is  $\mathbf{\Omega^U} = diag(\mathbf{\Omega^{U1}}, \mathbf{\Omega^{U2}}, \ldots, \mathbf{\Omega^{UJ}})$, where $\mathbf{\Omega^{Uj}} = {\mathbf{K^{Uj}}}^{-1}$. Similarly, $\mathbf{\Omega^S}$ is also a block diagonal with each block being a diagonal matrix $\mathbf{I}_{J^2}$. The marginal variational distribution of local latent function values $\mathbf{f^{Uj}}$ is a Gaussian with mean $\mathbf{m^{Uj}}$ and covariance $\mathbf{V^{Uj}}$, and that of dependent latent function values $\mathbf{f^{Sd}}$ is a Gaussian with mean $\mathbf{m^{Sd}}$ and covariance $\mathbf{V^{Sd}}$.
The  variational lower bound $L(\bm{\theta}, \bm{\gamma})$ requires computing an expectation of the log likelihood with respect to the variational distribution. However, the integral is intractable since the likelihood is a softmax function. So, we use Jensen's inequality to obtain a tractable lower bound to the expectation of log likelihood.
The variational lower bound $L(\bm{\theta}, \bm{\gamma})$ can be written as 
{\small \begin{eqnarray}
& &\hspace{-10mm}\frac{1}{2} \bigl( \sum_{j=1}^J (\log|\mathbf{V^{Uj}\Omega^{Uj}}| - tr(\mathbf{V^{Uj}\Omega^{Uj}}) - \mathbf{{m^{Uj}}^\top\Omega^{Uj} m^{Uj}})  \nonumber \\
& &\hspace{20mm}+ \sum_{d=1}^R (\log|\mathbf{V^{Sd}\Omega^{Sd}}|  - tr(\mathbf{V^{Sd}\Omega^{Sd}}) -  \mathbf{{m^{Sd}}^\top\Omega^{Sd} m^{Sd}})  \bigr) \nonumber \\
\label{eqn:lb2}
& &\hspace{-10mm}+\sum_{n=1}^{N} \sum_{l=1}^{L} \Bigl( {m}^{Uy_{nl}}_{nl} + \sum_{d=1}^R {m}^{Sd}_{(y^d_{nl},y_{nl})} - \log \bigl(  \sum_{q = 1}^J \exp ({m}^{Uj}_{nl} + \frac{1}{2} {V}^{Uj}_{(nl,nl)} \nonumber \\
& &\hspace{50mm}  + \sum_{d=1}^R {m}^{Sd}_{(y^d_{nl},q)} + \frac{1}{2} {V}^{Sd}_{((y^d_{nl},q),(y^d_{nl},q))}  ) \bigr) \Bigr).
\end{eqnarray}
}

The variational parameters $\bm{\gamma} = \{ \{\mathbf{m^{Uj}} \}_{j=1}^J, \{\mathbf{V^{Uj}} \}_{j=1}^J,$ $ \{\mathbf{m^{Sd}} \}_{d=1}^R, \{\mathbf{V^{Sd}} \}_{d=1}^R \}$ are estimated by maximizing the variational lower bound (\ref{eqn:lb2}). The  lower bound  is jointly concave with respect to all the variational parameters~\cite{boyd} and the optimum can be easily found using gradient based optimization techniques.

The variational parameters are estimated using a co-ordinate ascent approach. We repeatedly estimate each variational parameter while keeping the others fixed. The variational mean parameters $\mathbf{m^{Uj}}$ and $\mathbf{m^{Sd}}$ are estimated using gradient based approaches. The variational covariance matrices  $\mathbf{V^{Uj}}$ and $\mathbf{V^{Sd}}$  are  estimated under the positive semi-definite (p.s.d.) constraint. This can be done efficiently using the fixed point approach mentioned in \cite{khan12}. It is reported to converge faster than other VG approaches for GPs and is based on a concave objective function similar to (\ref{eqn:lb2}). The approach maintains the p.s.d. constraint on the covariance matrix and computes $\mathbf{V^{Uj}}$ by estimating only $\mathcal{O}(NL)$ variational parameters.
Estimation of $\mathbf{V^{Uj}}$ using the fixed point approach converges since (\ref{eqn:lb2}) is strictly concave with respect to $\mathbf{V^{Uj}}$.
The variational covariance matrix  $\mathbf{V^{Sd}}$ is diagonal since $\mathbf{\Omega^{Sd}}$ is diagonal. Hence, for computing a p.s.d. $\mathbf{V^{Sd}}$ we need to estimate only the diagonal elements of $\mathbf{V^{Sd}}$ under the element-wise non-negativity constraint. This can be done easily using gradient based methods.
The variational parameters $\bm{\gamma}$ are estimated for a particular set of hyper-parameters $\bm{\theta}$. The hyper-parameters $\bm{\theta}$ are also estimated by maximizing the lower bound (\ref{eqn:lb2}).  The variational parameters $\bm{\gamma}$ and the model parameters $\bm{\theta}$ are estimated alternately following a variational expectation maximization (EM) approach~\cite{pml}.
Algorithm~\ref{alg:vi} summarizes various steps involved in our approach.
\begin{algorithm}[tb]
   \caption{Model selection and learning in Gaussian process sequence labeling model}
   \label{alg:vi}
\begin{algorithmic}[1]
   \STATE {\bfseries Input:} Training data  ($\mathbf{X}$, $\mathbf{Y}$), dependency set $S$
   \STATE Initialize hyper-parameters $\bm{\theta}$, variational parameters $\bm{\gamma}$
   \REPEAT
   \REPEAT
   \FOR{$j=1$ {\bfseries to} $J$}
   \STATE Update $\mathbf{m^{Uj}}$ by maximizing (\ref{eqn:lb2}) \textit{w.r.t} $\mathbf{m^{Uj}}$
   \STATE Update $\mathbf{V^{Uj}}$ by maximizing (\ref{eqn:lb2}) \textit{w.r.t} $\mathbf{V^{Uj}}$
   \ENDFOR
   \FOR{$d=1$ {\bfseries to} $R$}
   \STATE Update $\mathbf{m^{Sd}}$ by maximizing (\ref{eqn:lb2}) \textit{w.r.t} $\mathbf{m^{Sd}}$
   \STATE Update $\mathbf{V^{Sd}}$ by maximizing (\ref{eqn:lb2}) \textit{w.r.t} $\mathbf{V^{Sd}}$
   \ENDFOR
   \UNTIL{relative increase in lower bound (\ref{eqn:lb2}) is small}
   \STATE Update $\bm{\theta}$ by maximizing (\ref{eqn:lb2}) \textit{w.r.t}  $\bm{\theta}$
   \UNTIL{relative increase in lower bound (\ref{eqn:lb2})  is small}
\STATE {\bfseries Return:} $\bm{\theta, \gamma}$
\end{algorithmic}
\end{algorithm}

The variational lower bound  (\ref{eqn:lb2}) is strictly concave with respect to each of the variational parameters. Hence, the estimation of variational parameters using co-ordinate ascent algorithm (inner loop) converges~\cite{nlp}. Convergence of EM for exponential family guarantees the convergence of Algorithm~\ref{alg:vi}. The overall computational complexity of  Algorithm~\ref{alg:vi} is dominated by the computation of $\mathbf{V^{Uj}}$. It takes $\mathcal{O}(JN^3L^3)$ time as it requires inversion of $J$ covariance matrices of size $NL \times NL$. The computational complexity for estimating $\mathbf{V^{Sd}}$ is $\mathcal{O}(RNLJ)$ and is negligible compared to the estimation of $\mathbf{V^{Uj}}$. Note that the computational complexity of the algorithm increases linearly with respect to the number of dependencies $R$. 

\section{Prediction}
\label{predgpsp}

We propose an iterative prediction algorithm which can effectively take into account the presence of multiple dependencies. 
The variational posterior distributions estimated using VG approximation $q(\mathbf{f^U}) = \prod_{j=1}^J q(\mathbf{f^{Uj}}) $ $= \prod_{j=1}^J \N(\mathbf{f^{Uj}}; \mathbf{m^{Uj}}, \mathbf{V^{Uj}})$ and 
$q(\mathbf{f^S}) =  \prod_{d=1}^R q(\mathbf{f^{Sd}})  = \prod_{d=1}^R$ $\N(\mathbf{f^{Sd}};\mathbf{m^{Sd}}, \mathbf{V^{Sd}})$  can be used to predict a  test output sequence  $\mathbf{y_*}$ given a test input sequence $\mathbf{\x_*}$. The predictive probability of assigning a label $y_{*l}$ to  a component of the output $\mathbf{y_*}$, given $\x_{*l}$ and rest of the labels $\mathbf{y}_{*}\backslash y_{*l}$ is 
\begin{eqnarray}
\label{eqn:pred}
  p(y_{*l} | \x_{*l},  \mathbf{y}_{*}\backslash y_{*l}) & =&  \int p(y_{*l} | \x_{*l}, \mathbf{y}_{*}\backslash y_{*l},\mathbf{f}_*) p(\mathbf{f}_*) d\mathbf{f}_* \nonumber \\
 & =& \int \frac{\exp(f^{Uy_{*l}}_{*l} + \sum_{d =1}^R f^{Sd}_{*}(y^d_{*l},y_{*l}))}{\sum_{y_{*l} = 1}^J \exp(f^{Uy_{*l}}_{*l} + \sum_{d=1}^R f^{Sd}_{*}(y^d_{nl},y_{nl}))} \nonumber \\
 & & \hspace{6mm} \{p(f^{Uj}_{*l})\}_{j=1}^J \{p(f^{Sd}_*)\}_{d=1}^R \{df^{Uj}_{*l}\}_{j=1}^J \{df^{Sd}_*\}_{d=1}^R
\end{eqnarray}
where $p(\mathbf{f}_*)$ denotes the predictive distribution of all the latent function values for the test input $\mathbf{\x_*}$.
In (\ref{eqn:pred}), $p(f^{Uj}_{*l})$ represents the predictive distribution of the local latent function $j$ for a test input component $\x_{*l}$. This is Gaussian with mean $m^{Uj}_{*l}$ and variance $v^{Uj}_{*l}$ where,
\begin{eqnarray}
& & \hspace{-10mm} m^{Uj}_{*l} = {\mathbf{K}^\mathbf{Uj}_{*l}}^{\top}\mathbf{\Omega^{Uj}}\mathbf{m^{Uj}} \quad \mbox{    and}\nonumber \\
& & \hspace{-10mm} v^{Uj}_{*l}  = K^{Uj}_{*l,*l} - {\mathbf{K}^\mathbf{Uj}_{*l}}^\top(\mathbf{\Omega^{Uj}} - \mathbf{\Omega^{Uj}}\mathbf{V^{Uj}}\mathbf{\Omega^{Uj}}){\mathbf{K}^\mathbf{Uj}_{*l}}. \nonumber
\end{eqnarray}
Here, $\mathbf{K}^\mathbf{Uj}_{*l}$ is an $NL$ dimensional vector obtained from the kernel evaluations for the label $j$ between the test input data component $\x_{*l}$ and the training data $\mathbf{X}$ and $K^{Uj}_{*l,*l}$ represents the kernel evaluation of the test data input component $\x_{*l}$ with itself. $\mathbf{f^{Sd}}$ is independent of the test data input and the predictive distribution $p(\mathbf{f_{*}^{Sd}})$ is the same as  $p(\mathbf{f^{Sd}})$.  This is a Gaussian with mean $\mathbf{m^{Sd}}$ and covariance $\mathbf{V^{Sd}}$. The computation of the expected value of softmax with respect to the latent functions (\ref{eqn:pred}) is intractable. Instead we compute softmax of the expected value of the latent functions and compute a normalized probabilistic score.
We refine the normalized score to take into account the uncertainty in true labels associated with the dependencies and compute the refined normalized score ($RNS$) as
\begin{eqnarray}
\label{eqn:rns}
& &RNS(y_{*l}, \x_{*l}) = \frac{\exp(m^{Uy_{*l}}_{*l} + \frac{1}{2} v^{Uy_{*l}}_{*l} + \sum_{d=1}^R \mathbb{E}_{y^d_{*l}} [g^{d}(y^d_{*l},y_{*l})])}{ \sum_{q = 1}^J \exp(m^{Uj}_{*l} + \frac{1}{2} v^{Uj}_{*l}  + \sum_{d=1}^R \mathbb{E}_{y^d_{*l}} [g^{d}(y^d_{*l},q) ])} \nonumber
\end{eqnarray}
Here, $g^{d}(y^d,y) = \mathbf{m}^\mathbf{Sd}_{(y^d,y)} + \frac{1}{2} \mathbf{V}^\mathbf{Sd}_{((y^d,y),(y^d,y))}$ determines the contribution of the label $y^d$ of dependency $d$ in predicting the output label $y$. $RNS$ considers an expected value over all the possible labelings associated with a dependency $d$.  The expectation is computed using the $RNS$ value associated with the labels $y^d_{*l}$ for the input $x^d_{*l}$,  that is,  $\mathbb{E}_{y^d_{*l}}[\cdot] = \sum_{y^d_{*l} = 1}^J RNS(y^d_{*l}, x^d_{*l}) [\cdot]$.

We provide an iterative approach to estimate the labels of a test output in Algorithm~\ref{alg:pred}.
An initial $RNS$ value is computed without considering the dependencies. We iteratively refine the $RNS$ value using the previously computed $RNS$ value by taking into account the dependencies. The process is continued until convergence. The final $RNS$ value is used to make prediction separately for each output component by assigning labels with the maximum $RNS$ value. The computational complexity of Algorithm~\ref{alg:pred} is $\mathcal{O}(J^2RL)$ and is same as that of Viterbi algorithm~\cite{hmm} for a single dependency case. The convergence of Algorithm~\ref{alg:pred} follows from the analysis presented in \cite{Li13} for a similar fixed point algorithm. The algorithm is found to converge in a few iterations in our experiments.

\begin{algorithm}[t]
   \caption{Prediction in Gaussian process sequence labeling model}
   \label{alg:pred}
\begin{algorithmic}[1]
   \STATE {\bfseries Input:} Test data $\mathbf{x}_* = (\x_{*1}, \ldots, \x_{*L})$, posterior mean $\{\mathbf{m^{Uj}}\}_{j=1}^J$ and $\{\mathbf{m^{Sd}}\}_{d=1}^R$ and posterior covariance $\{\mathbf{V^{Uj}}\}_{j=1}^J$ and $\{\mathbf{V^{Sd}}\}_{d=1}^R$
   \STATE Obtain predictive means $\{\{m^{Uj}_{*l}\}_{j=1}^J\}_{l=1}^L$, and variances $\{\{v^{Uj}_{*l}\}_{j=1}^J\}_{l=1}^L$
   \STATE {\bfseries Initialize :} $RNS^0(y_{*l}, \x_{*l}) =  \frac{\exp(m^{Uy_{*l}}_{*l} + \frac{1}{2} v^{Uy_{*l}}_{*l} )}{ \sum_{j = 1}^J \exp(m^{Uj}_{*l} + \frac{1}{2} v^{Uj}_{*l})}$ $ \, \forall \, y_{*l} = 1, \ldots, J , \,\forall \,l=1\ldots, L$
   \STATE {\bfseries Initialize :} $t = 0$
   \REPEAT
   \STATE $t = t+1$
   \FOR{$l=1$ {\bfseries to} $L$}
   \FOR{$y_{*l} = 1$ {\bfseries to} $J$}
 \STATE $RNS^t(y_{*l}, \x_{*l})=\frac{\exp(m^{Uy_{*l}}_{*l} + \frac{1}{2} v^{Uy_{*l}}_{*l} + \sum_{d=1}^R \mathbb{E}_{y^d_{*l}} [g^{d}(y^d_{*l},y_{*l})])}{ \sum_{j = 1}^J \exp(m^{Uj}_{*l} + \frac{1}{2} v^{Uj}_{*l}  + \sum_{d=1}^R \mathbb{E}_{y^d_{*l}} [g^{d}(y^d_{*l},q) ])}$
 \STATE \mbox{where }$\mathbb{E}_{y^d_{*l}} [\cdot] = \sum_{y^d_{*l} = 1}^J RNS^{t-1}(y^d_{*l}, x^d_{*l}) [\cdot]$
   \ENDFOR
   \ENDFOR
   \UNTIL{ change in $RNS^t$ \textit{w.r.t} $RNS^{t-1}$ is small}
\STATE
$(\hat{y}_{*1}, \ldots, \hat{y}_{*L}) = (\argmax_{y_{*1}} RNS^t(y_{*1}, \x_{*1}), \ldots,$ \\  $\qquad \qquad \qquad \qquad \quad \argmax_{y_{*L}} RNS^t(y_{*L}, \x_{*L}))$
\STATE {\bfseries Return:} $(\hat{y}_{*1}, \ldots, \hat{y}_{*L})$
\end{algorithmic}
\end{algorithm}

\section{Experimental Results}
\label{exptgpsp}

We conduct experiments to study the generalization performance of the proposed Gaussian Process Sequence labeling (GPSL) model.  We use the sequence labeling problems in natural language processing to study the behavior of the proposed approach.  Although the proposed approach is general and can handle dependencies of any length, we consider three different models of the proposed approach in our experiments. The first model, GPSL1, assumes that the current label depends only on the previous label. The second model, GPSL2, assumes that the current label depends both on the previous and the next label in the sequence. The third model, GPSL4, assumes that the current label depends on the previous two labels and the next two labels.

We consider four sequence labeling problems in natural language processing to study the performance of the proposed approach. The datasets for all these problems are obtained from the CRF++\footnote{Available at http://crfpp.googlecode.com/svn/trunk/doc/index.html} toolbox. We provide a brief description of the tasks in each of these data sets.  \\
{\bf Base NP :} We need to identify noun phrases in a sentence. The starting word in the noun phrase is given a label $B$, while the words inside the noun phrase are given a label $I$. All the other words are given a label $O$. The task here is to  assign each word with a label from the set $\{B,I,O\}$. \\
{\bf Chunking :} Shallow parsing or chunking identifies constituents in a sentence such as noun phrase, verb phrase etc. Here, each word in a sentence is labeled as belonging to verb phrase, noun phrase etc. In the  $Chunking$ dataset, words are assigned a label from a set of size $14$. \\
{\bf Segmentation :} Segmentation is the process of finding meaningful segments in a text such as words, sentences etc. We consider a word segmentation problem where the words are  identified from a Chinese sentence. The $Segmentation$ data set assigns each unit in the sentence a label denoting whether it is beginning of a word ($B$) or inside a word ($I$). The task is to assign either of these two labels to each unit in a sentence.\\
{\bf Japanese NE :} We need to perform Named Entity Recognition (NER) where the task is to identify whether the words in  a sentence denote a named entity such as person, place, time etc. We use the $Japanese NE$ dataset where the Japanese words are assigned one of $17$ different named entities. \\

In all these data sets except $Segmentation$, a sentence is considered as an input and words in the sentence as input components. In $Segmentation$, every  alphabet is considered
as an input component.  The features for  each input component are extracted using the template files provided in the CRF++ package. The properties of all the  data sets are summarized in Table~\ref{tab:gpsl}. It mentions the number of sentences ($N$) used for training and testing. The effective sample size ($NL$) for the GPSL models is obtained by multiplying this quantity by average sentence length which increases the data size by an order of magnitude.

We compare the performance of the proposed approach with popular sequence labeling approaches, structural SVM (SSVM)~\cite{shevade11}\footnote{Code available at  http://drona.csa.iisc.ernet.in/$\sim$shirish/structsvm\_sdm.html}, conditional random field (CRF)~\cite{bottou10}\footnote{Code available at  http://leon.bottou.org/projects/sgd\#stochastic\_gradient\_descent\_version\_2}, and GPstruct~\cite{bratiers13}\footnote{Code available at  https://github.com/sebastien-bratieres/pygpstruct}. All the models used a linear kernel. GPstruct experiments are run for 100000 elliptical slice sampling steps. The performance is measured in terms of average Hamming loss over all the test data points. The Hamming loss between the actual test output $\mathbf{y}_*$ and the predicted test output $\mathbf{\hat{y}}_*$ is given by $Loss(\mathbf{y}_*,\mathbf{\hat{y}}_*) = \sum_{l=1}^L \mathbb{I}(y_{*l} \neq \hat{y}_{*l})$, where $\mathbb{I}(\cdot)$ is the indicator function.   Table \ref{tab:gpsl} compares the performance (percentage 
of the average Hamming loss) of various approaches on the four sequence labeling problems. The GPSL models, SSVM, CRF and GPstruct are run over $10$ independent partitions of the data set\footnote{The train and test set partitions  are different from those used by \cite{bratiers13}.} and a mean of the Hamming loss over all the partitions along with the standard deviation are reported in Table~\ref{tab:gpsl}.

\begin{table*}[t]
\caption{ Properties of the sequence labeling data sets and a comparison of the performance of various models on these data sets. The approaches GPSL1, GPSL2, GPSL4, SSVM, CRF and GPstruct are compared using average Hamming loss (in percentage).  The numbers in bold face style indicate the best results among these approaches. `$\star$' and `$\dagger$' denote if the performance of a method is significantly different from the best performing method and GPstruct repectively, according to paired t-test with 5\% significance level.}
\label{tab:gpsl} 
\centering{
\begin{tabularx}{\textwidth}{|C|C|C|C|C|}
\hline
 & Base NP & Chunking & Segmentation &  Japanese NE \\
\hline	
\#labels  & 3 & 14 & 2  & 17 \\
\#features  & 6438 & 29764 & 1386 & 102,799 \\
training/ test sentences & 150/150 & 50/50 & 20/16 & 50/50  \\
\hline
GPSL1 & 5.73$\pm$0.98$\star$  &    13.02$\pm$1.87$\star$   &   \bf{23.45$\pm$2.96} &  8.26$\pm$2.63$\star$  \\
GPSL2 & 5.55$\pm$0.92$\star$ &  12.69$\pm$1.69$\star$   & 23.51$\pm$2.93 &  7.86$\pm$2.45 $\star$  \\
GPSL4 & 5.54$\pm$0.94$\star$  &  12.70$\pm$1.79$\star$  &  23.53$\pm$2.85 &  7.82$\pm$2.56 $\star$ \\
CRF  & 5.21$\pm$0.84$\dagger$  &  11.76$\pm$1.73$\star$$\dagger$    &    24.10$\pm$3.49$\star$$\dagger$     &  7.76$\pm$2.80 $\star$ \\
SSVM & \bf{5.19$\pm$0.91}$\dagger$ & \bf{10.71$\pm$1.49}$\dagger$  &  23.46$\pm$3.45 &  \bf{6.17$\pm$2.60}$\dagger$ \\
GPstruct & 5.66$\pm$0.93$\star$ & 12.56$\pm$1.82$\star$  & 23.55$\pm$2.90  &  7.79$\pm$2.92 $\star$ \\
\hline
\end{tabularx}
}
\end{table*}

The reported results show that the GPSL models with multiple dependencies performed better than GPstruct on  $BaseNP$ and $Segmentation$. In the other two data sets, GPSL models came close to GPstruct. We find that increasing the number of dependencies helped to improve the performance in general except for the $Segmentation$ data set. This is due to the difference in nature of the sequence labeling task involved in segmentation. For other data sets, the GPSL model which considered
both the previous and next label (GPSL2) gave a better performance. The performance of the GPSL model which considered the previous and the next 2  labels (GPSL4) improved only marginally or worsened compared to GPSL2 on these data sets. We note that increasing the  number of dependencies beyond four did not bring any improvement in performance for the sequence labeling data sets that we have considered. Overall, the performance of the SSVM is found to be better than other approaches in these sequence labeling data sets. However, GPSL models have the advantage of being Bayesian and can provide a confidence over label predictions which is useful for many NLP tasks.

\subsection{ Runtime performance of the GPSL models}

The proposed GPSL models are implemented in Matlab. The GPSL Matlab programs are run on a 3.2 GHz Intel processor with 4GB of shared main memory under Linux.  The SSVM approach is implemented in C, the CRF approach is coded in C++ and the GPStruct approach is in Python. Since the implementation languages differ, it is unfair to make a runtime comparison of various approaches.  Table~\ref{table:GPSLruntime}  compares the average runtime (in seconds) for training various GPSL models and GPstruct on the sequence labeling data sets. We find that the GPSL models are an order of magnitude faster than GPStruct. We also find that increasing the dependencies resulted in only a slight increase in runtime.


\begin{table}[t]
\renewcommand{\arraystretch}{1}
\renewcommand\tabcolsep{1pt}
\caption{\small{Comparison of  average running time (seconds) of various GPSL models and GPstruct} }
\label{table:GPSLruntime}
\centering{
\begin{tabular}{|c|c|c|c|c|}
\hline
Data  &  GPSL1  &  GPSL2      & GPSL4 & GPstruct \\
\hline
Segmentation  & 17.13   &  19.64  & 22.83 & 3.82e+03 \\
Chunking  & 1.09e+03  &  1.35e+03 &  1.71e+03    & 4.56e+04 \\
Base NP  & 6.01e+03 &   6.69e+03  &   7.25e+03 & 7.54e+04 \\
Japanese NE &   1.24e+03  &  1.56e+03   & 1.93e+03 &  4.92e+04  \\
\hline
\end{tabular}}
\end{table}

\begin{table}[t]
\renewcommand{\arraystretch}{1}
\renewcommand\tabcolsep{1pt}
\caption{\small{Comparison of the prediction algorithms using GPSL1 model} }
\label{table:GPSLprediction}
\centering{
\begin{tabular}{|c|c|c|c|c|c|c|}
\hline
 & \multicolumn{2}{c|}{average Hamming loss} & paired t-test & \multicolumn{2}{c|}{average runtime (seconds)} & average iterations \\
\hline
Data  &  Algorithm~\ref{alg:pred}  & Viterbi  & t-value   & Algorithm~\ref{alg:pred} & Viterbi & Algorithm~\ref{alg:pred} \\
\hline
Segmentation  & 23.45 & 24.26  & 3.8183 &  0.1227 & 0.0856 & 5 \\
Chunking  & 13.02   &  13.69 & 3.6421  & 0.2491 & 0.2628 & 5\\
Base NP  & 5.73   &  5.75 & 0.3162  & 0.5207 & 0.5338  & 4\\
Japanese NE  & 8.26   &  8.84 & 2.475  & 0.3661 & 0.5653 & 3\\
\hline
\end{tabular}}
\end{table}

\begin{figure}[t]
\begin{center}
\subfloat[\scriptsize Base NP]{
\includegraphics[scale=0.28]{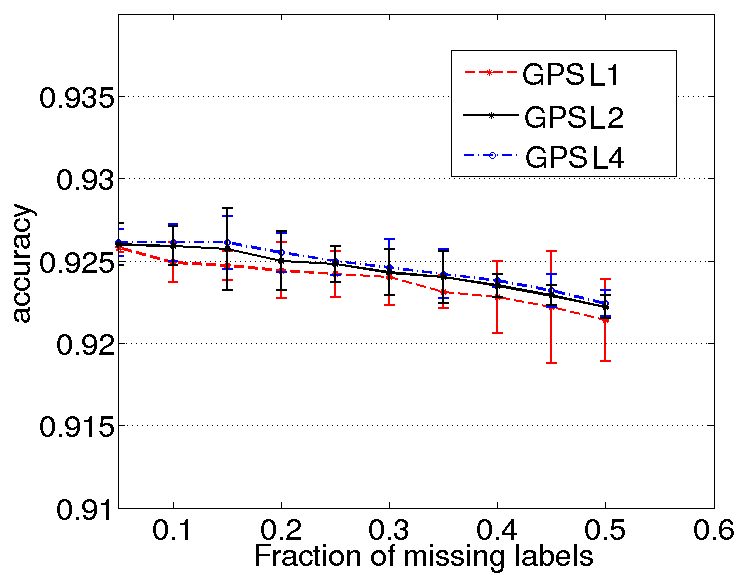}
}
\subfloat[\scriptsize Chunking]{
\includegraphics[scale=0.28]{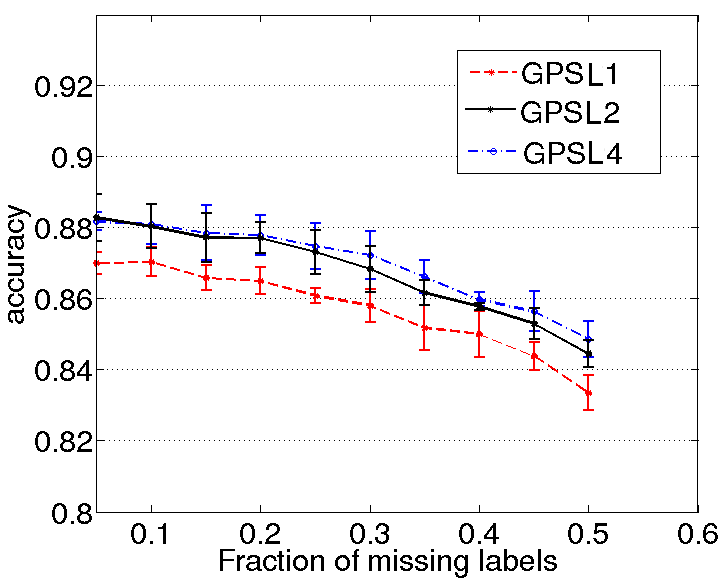}
}
\\
\subfloat[\scriptsize Segmentation]{
\includegraphics[scale=0.28]{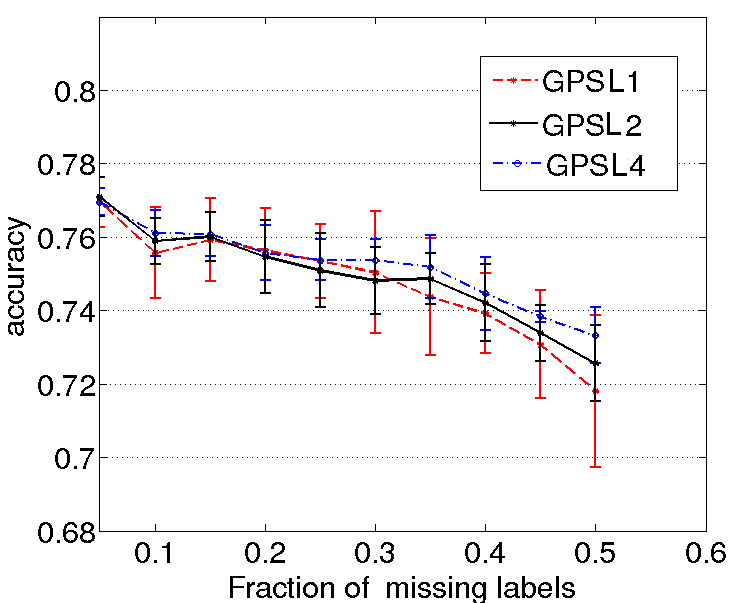}
}
\subfloat[\scriptsize Japanese NE]{
\includegraphics[scale=0.28]{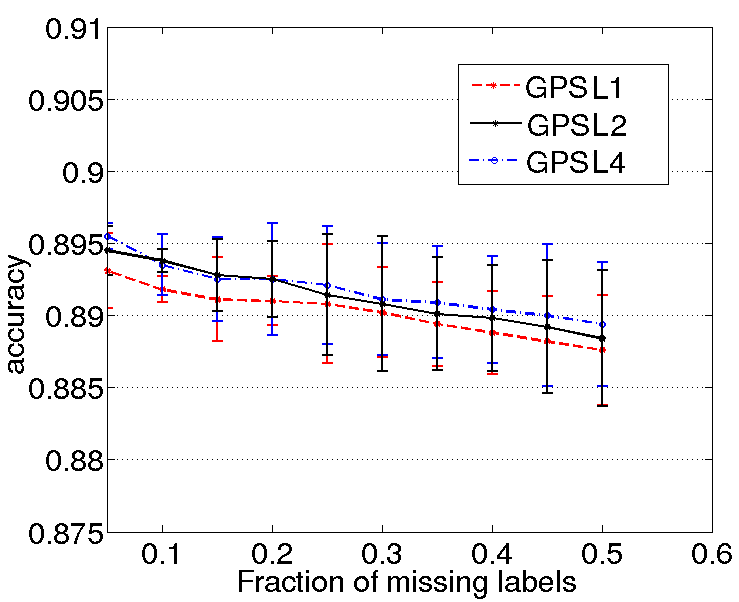}
}
\caption{\small Variation in accuracy as the fraction of missing labels is varied from $0.05$ to $0.5$ }
\label{fig:exptml}
\end{center}
\end{figure}

\subsection{ Experiments with the Prediction algorithm}

We conducted experiments to study the performance of  Algorithm~\ref{alg:pred} used to make prediction. The algorithm is compared with the commonly used Viterbi algorithm~\cite{hmm} for the sequence labeling task. Viterbi algorithm consists of a forward phase which calculates the best value attained at the end of the sequence and a backward phase which finds the sequence of labels that lead to it.  It is useful only for the setting where one considers a dependency with the previous label. Therefore, we study how the performance of the GPSL1 model differs when Viterbi algorithm is used for prediction instead of the proposed algorithm. We consider an implementation of the Viterbi algorithm provided by the UGM toolkit~\cite{ugm}. Table~\ref{table:GPSLprediction} compares the predictive and runtime performance of the two algorithms. We observe that Algorithm~\ref{alg:pred} gave a better predictive and runtime performance than the Viterbi algorithm. The predictive performance of Algorithm~\ref{alg:pred} is significantly better than Viterbi on $Segmentation$, $Chunking$ and $Japanese NE$. The t-values calculated using paired t-test on these data sets are found to be greater than the critical value of $2.262$ for a level of significance $0.05$ and $9$ degrees of freedom. We also observed that Algorithm~\ref{alg:pred} converged in 3-5 iterations on an average.
 
\subsection{ Experiments with Missing Labels}

In many sequence labeling tasks in NLP, the labels of some of the output components might be missing in the training data set. This is  common when crowd sourcing techniques are employed to obtain the labels.
Sequence labeling approaches such as SSVM and CRF are not readily applicable to data sets with missing labels.
GPSL models are useful to learn from the data sets with missing labels due to their ability to capture larger dependencies.  We learn the GPSL models from the sequence labeling data sets with some  fraction of the labels missing. We vary the fraction of  missing labels and study how the performance of our model varies with respect to missing labels.  Figure \ref{fig:exptml} provides the variation in performance  of various GPSL models as we vary the fraction of  missing labels. The performance is measured in terms of accuracy which is obtained by subtracting the average Hamming loss from $1$.
We find that the performance of the GPSL models  does not significantly degrade as the fraction of the missing labels increases. Figure \ref{fig:exptml} shows that GPSL4 which uses the previous  and the next 2 labels provides a better
performance than the other GPSL models. GPSL4 learns a better model by considering a larger neighborhood information and is useful to handle  data sets with missing labels.

\section{Conclusion}
\label{congpsp}
We proposed a novel Gaussian Process approach to perform sequence labeling based on pseudo-likelihood approximation. The use of  pseudo-likelihood  enabled the model to capture multiple dependencies without becoming computationally intractable. The approach used a faster inference scheme based on variational inference. We also proposed an approach to perform prediction which  makes use of the information from the neighboring labels. The proposed approach is useful for a wide range of sequence labeling problems arising in natural language processing. Experimental results showed that GPSL models, which capture multiple dependencies, are useful in sequence labeling problems. The ability to capture multiple dependencies makes them effective in handling data sets with missing labels.

\bibliography{VBGPSP}

\begin{thebibliography}{10}
\providecommand{\url}[1]{\texttt{#1}}
\providecommand{\urlprefix}{URL }

\bibitem{altun04}
Altun, Y., Hofmann, T., Smola, A.J.: {Gaussian Process Classification for
  Segmenting and Annotating Sequences}. In: ICML (2004)

\bibitem{shevade11}
Balamurugan, P., Shevade, S., Sundararajan, S., Keerthi, S.: {A Sequential Dual
  Method for Structural SVMs.} In: SDM. pp. 223--234 (2011)

\bibitem{nlp}
Bertsekas, D.P.: {Nonlinear Programming}. Athena Scientific (1999)

\bibitem{besag1975}
Besag, J.: {Statistical analysis of non-lattice data}. The Statistician  24,
  179--195 (1975)

\bibitem{bottou10}
Bottou, L.: {Large-Scale Machine Learning with Stochastic Gradient Descent.}
  In: COMPSTAT (2010)

\bibitem{boyd}
Boyd, S., Vandenberghe, L.: {Convex Optimization}. Cambridge University Press
  (2004)

\bibitem{bratiers13}
Bratieres, S., Quadrianto, N., Ghahramani, Z.: {Bayesian Structured Prediction
  Using Gaussian Processes}. IEEE transactions on Pattern Analysis and Machine
  Intelligence  (2014)

\bibitem{bratiers14}
Bratieres, S., Quadrianto, N., Nowozin, S., Ghahramani, Z.: {Scalable Gaussian
  Process Structured Prediction for Grid Factor Graph Applications}. ICML
  (2014)

\bibitem{chai12}
Chai, K.M.A.: {Variational Multinomial Logit Gaussian Process}. J. Mach. Learn.
  Res.  13 (2012)

\bibitem{girolami06}
Girolami, M., Rogers, S.: {Variational Bayesian Multinomial Probit Regression
  with Gaussian Process Priors}. Neural Computation  18(8),  1790--1817 (2006)

\bibitem{heckerman01}
Heckerman, D., Chickering, D.M., Meek, C., Rounthwaite, R., Kadie, C.:
  {Dependency Networks for Inference, Collaborative Filtering, and Data
  Visualization}. J. Mach. Learn. Res.  1,  49--75 (2001)

\bibitem{khan12}
Khan, M.E., Mohamed, S., Murphy, K.P.: {Fast Bayesian Inference for
  Non-Conjugate Gaussian Process Regression}. In: NIPS. pp. 3149--3157 (2012)

\bibitem{crf}
Lafferty, J.D., McCallum, A., Pereira, F.C.N.: {Conditional Random Fields:
  Probabilistic Models for Segmenting and Labeling Sequence Data}. In: ICML.
  pp. 282--289 (2001)

\bibitem{kcrf}
Lafferty, J.D., Zhu, X., Liu, Y.: {Kernel Conditional Random Fields:
  Representation and Clique Selection.} In: ICML (2004)

\bibitem{Li13}
Li, Q., Wang, J., Wipf, D.P., Tu, Z.: {Fixed-Point Model For Structured
  Labeling}. In: ICML. pp. 214--221 (2013)

\bibitem{pml}
Murphy, K.P.: {Machine learning: A Probabilistic Perspective}. The MIT Press
  (2012)

\bibitem{lsp}
Noah, A.S.: {Linguistic Structure Prediction }. Morgan and Claypool (2011)

\bibitem{opper09}
Opper, M., Archambeau, C.: {The Variational Gaussian Approximation Revisited}.
  Neural Computation  21,  786--792 (2009)

\bibitem{bcrf}
Qi, Y., Szummer, M., Minka, T.P.: {Bayesian conditional random fields}. Proc.
  AISTATS  (2005)

\bibitem{hmm}
Rabiner, L.R.: {A Tutorial on Hidden Markov Models and Selected Applications in
  Speech Recognition}. Proceedings of the IEEE  77(2),  257--286 (1989)

\bibitem{ras05}
Rasmussen, C.E., Williams, C.K.I.: {Gaussian Processes for Machine Learning
  (Adaptive Computation and Machine Learning)}. MIT Press (2005)

\bibitem{ugm}
Schmidt., M.: {UGM: A Matlab toolbox for probabilistic undirected graphical
  models.} (2007), \url{http://www.cs.ubc.ca/~schmidtm/Software/UGM.html}

\bibitem{sutton07}
Sutton, C., McCallum, A.: {Piecewise Pseudolikelihood for Efficient Training of
  Conditional Random Fields}. pp. 863--870. ICML (2007)

\bibitem{toutanova03}
Toutanova, K., Klein, D., Manning, C.D., Singer, Y.: {Feature-Rich
  Part-of-Speech Tagging with a Cyclic Dependency Network}. In: HLT-NAACL. pp.
  252--259 (2003)

\bibitem{structSVM}
Tsochantaridis, I., Joachims, T., Hofmann, T., Altun, Y.: {Large Margin Methods
  for Structured and Interdependent Output Variables}. J. Mach. Learn. Res.  6,
   1453--1484 (2005)

\bibitem{williams98}
Williams, C.K.I., Barber, D.: {Bayesian Classification with Gaussian
  Processes}. IEEE Transactions on Pattern Analysis and Machine Intelligence
  20(12),  1342--1351 (1998)

\end{thebibliography}
\bibliographystyle{splncs03}

\end{document}